\documentclass[10pt,twocolumn,letterpaper]{article}

\usepackage{iccv}              
\usepackage{xcolor}
\usepackage[accsupp]{axessibility}  

\definecolor{iccvblue}{rgb}{0.21,0.49,0.74}
\usepackage[pagebackref,breaklinks,colorlinks,allcolors=iccvblue]{hyperref}

\def\paperID{11} 
\def\confName{ICCV}
\def\confYear{2025}

\title{$\text{WS}^2$: Weakly Supervised Segmentation using Before-After \\ Supervision in Waste Sorting}

\author{%
  Andrea Marelli\textsuperscript{1} 
  \quad Alberto Foresti\textsuperscript{2} 
  \quad Leonardo Pesce\textsuperscript{1}
  \quad Giacomo Boracchi\textsuperscript{1}
  \quad Mario Grosso\textsuperscript{1}\\
  {\small\textsuperscript{1} Politecnico di Milano, Italy \quad \textsuperscript{2} EURECOM, France}\\
  \tt\small andrea.marelli@polimi.it, alberto.foresti@eurecom.fr, leonardo.pesce@mail.polimi.it, \\
  \tt\small giacomo.boracchi@polimi.it, mario.grosso@polimi.it
}

\begin{document}
\maketitle
\begin{abstract}
In industrial quality control, to visually recognize unwanted items within a moving heterogeneous stream, human operators are often still indispensable. Waste-sorting stands as a significant example, where operators on multiple conveyor belts manually remove unwanted objects to select specific materials. To automate this recognition problem, computer vision systems offer great potential in accurately identifying and segmenting unwanted items in such settings. Unfortunately, considering the multitude and the variety of sorting tasks, fully supervised approaches are not a viable option to address this challange, as they require extensive labeling efforts. Surprisingly, weakly supervised alternatives that leverage the implicit supervision naturally provided by the operator in his removal action are relatively unexplored. In this paper, we define the concept of Before-After Supervision, illustrating how to train a segmentation network by leveraging only the visual differences between images acquired \textit{before} and \textit{after} the operator. To promote research in this direction, we introduce $\text{WS}^2$ (Weakly Supervised segmentation for Waste-Sorting), the first multiview dataset consisting of more than 11 000 high-resolution video frames captured on top of a conveyor belt, including "before" and "after" images. We also present a robust end-to-end pipeline, used to benchmark several state-of-the-art weakly supervised segmentation methods on $\text{WS}^2$. \footnote{The $\text{WS}^2$ dataset is publicly available for download at \url{https://zenodo.org/records/14793518}, all the details are reported in the supplementary material.}
\end{abstract}
    
\begin{figure}[tbp]
\centerline{\includegraphics[width=0.45\textwidth]{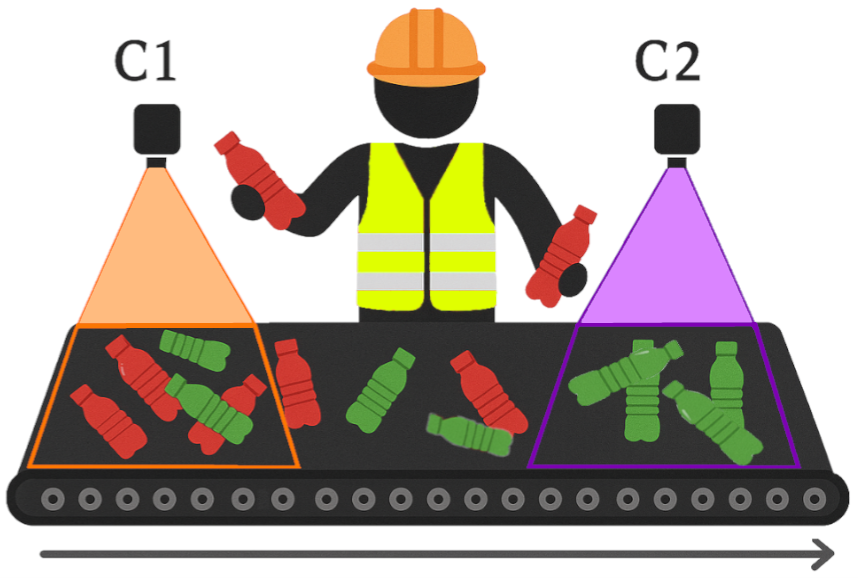}}
\caption{In a waste sorting plant, two cameras, $C_1$ and $C_2$, are placed along a conveyor belt where a human operator manually removes \textit{\textcolor{red}{unwanted}} objects (\textcolor{red}{red}) from a mixed waste stream, leaving on the belt only \textit{\textcolor{green}{wanted}} ones (\textcolor{green}{green}). $C_1$ captures the belt section \textit{before} the HO’s intervention, while $C_2$ captures the section \textit{after}, where only \textit{\textcolor{green}{wanted}} objects remain. We aim to train, without any pixel-wise annotation nor additional supervision, a binary segmentation model for \textit{\textcolor{green}{wanted}}/\textit{\textcolor{red}{unwanted}} objects.} 
\label{fig:camera_scenario}
\end{figure}

\begin{figure*}[t]
\begin{minipage}[b]{.30\textwidth}
  \centering
  \includegraphics[width=\textwidth]{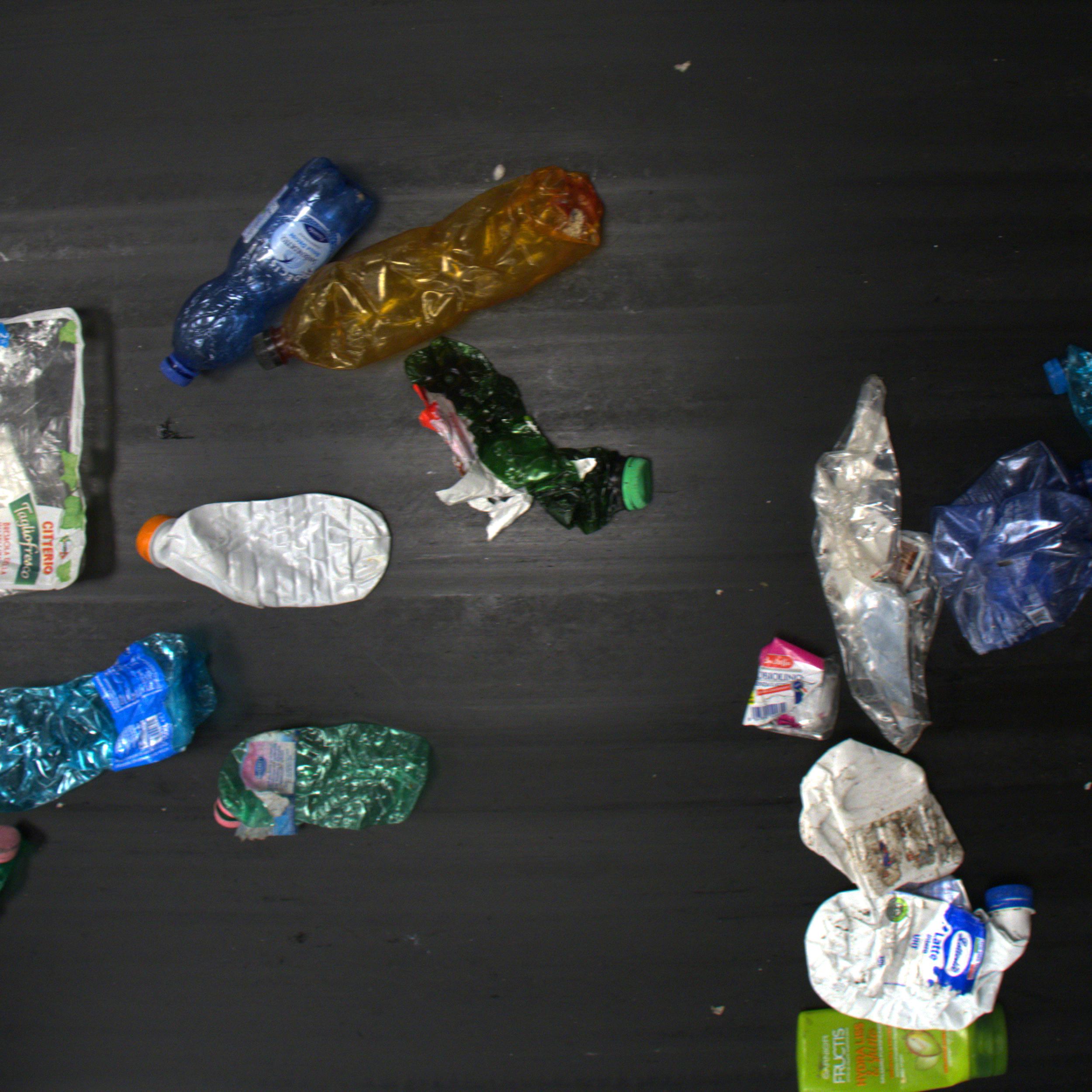}\\
  (a)
\end{minipage}
\hfill
\begin{minipage}[b]{.30\textwidth}
  \centering
  \includegraphics[width=\textwidth]{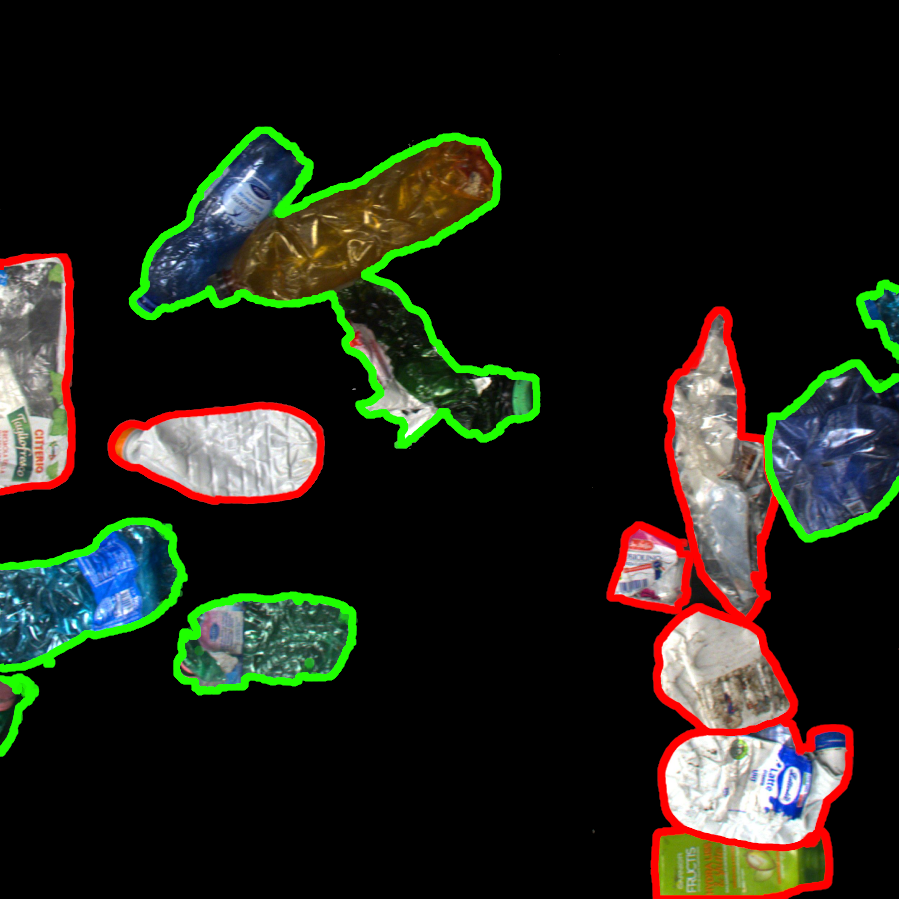}\\
  (b)
\end{minipage}
\hfill
\begin{minipage}[b]{.30\textwidth}
  \centering
  \includegraphics[width=\textwidth]{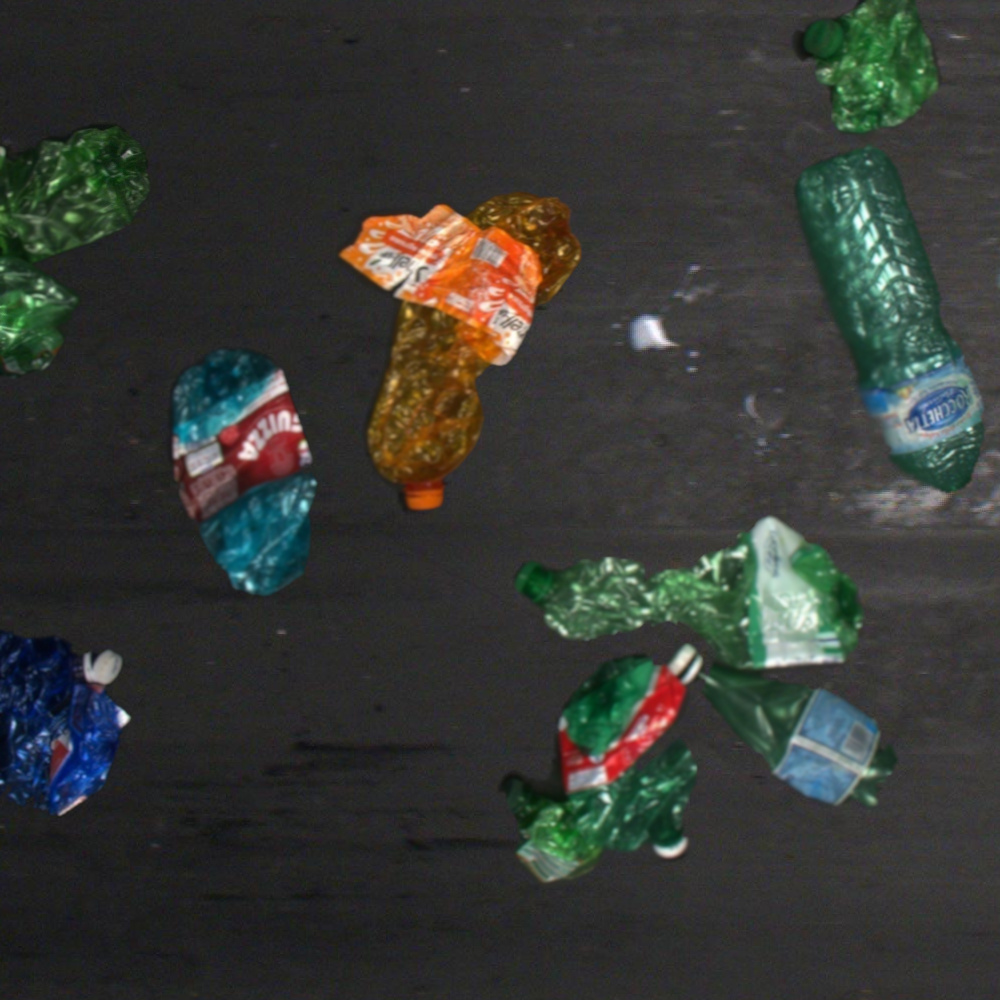}\\
  (c)
\end{minipage}
\caption{
    (a) “\textit{Before}” frame showing both \textit{wanted} and \textit{unwanted} waste items on the conveyor belt. 
    (b) Annotation overlay: green outlines mark the \textit{wanted} items to keep (semi-transparent PET bottles), while red outlines indicate \textit{unwanted} waste to be removed (all other plastic materials).
    (c) “\textit{After}” frame in which only the \textit{wanted} items remain. 
    The absence of the \textit{unwanted} waste, present only in the “\textit{before}” frames, serves as supervision in our WS segmentation approach, enabling the training of a network able to segment the \textit{unwanted} items.
}
\label{fig:retain_and_remove}
\end{figure*}

\section{Introduction}
\label{sec:intro}

Despite the increasing automation in multiple industrial sectors, Human Operators (HOs) are still strictly necessary for numerous tasks. Many quality control processes, ranging from food to pharmaceutical lines, rely on HOs meticulously monitoring a continuous flow of items, identifying and manually removing anomalies or imperfections based on each item’s visual appearance. This principle is reflected in modern recycling plants. Here, waste is first coarsely sorted by NIR-based automatic machines, then ends up on multiple separate conveyor belts where HOs manually remove undesired items from the stream, leaving on each dedicated belt only the objects of a specific material based on visual cues. Figure~\ref{fig:camera_scenario} illustrates this process: the HO discards the \textit{unwanted} objects, leaving on the belt only the \textit{wanted} ones. Supporting this activity with automatic tools would improve both the efficacy and efficiency of sorting plants, reducing the risks of injuries and alleviating stressful working conditions for HOs \cite{ijerph120201216}. With the advent of deep learning, computer vision systems have emerged as promising solutions to automating sorting tasks. In fact, a semantic segmentation network can be trained to analyze images or videos capturing the stream running on a belt and segment the \textit{unwanted} items, supporting humans in their operation and paving the way for the system's automation. 

In spite of substantial research on industrial waste sorting segmentation, existing efforts have been mainly based on datasets designed for Fully Supervised (FS) learning (Table~\ref{tab:db_table}). Unfortunately, gathering extensive manual annotations in the form of pixel-level annotated segmentation masks can be very time-consuming. Considering the fine-grained sorting demands of modern recycling plants, a dedicated detection model would be required for each specific line of work. Therefore, training each specific model in a FS fashion would require a large amount of data to be annotated for every belt, resulting in prohibitive costs.


In this paper, we investigate an alternative learning paradigm for segmenting \textit{unwanted} items in sorting tasks, without the need to manually annotate the large quantity of ground-truth masks required to train a distinct segmentation network for each individual line. Our key insight lies in the fact that, comparing images of the belt sections captured \textit{before} and \textit{after} the HO's removal operation (Figure~\ref{fig:retain_and_remove}), it is possible to learn the supervisory signals for the segmentation task that is already implicit in the HO's activity. In fact, by removing the \textit{unwanted} objects from the stream, the operator effectively indicates which items should be segmented without any manual annotations. We define this special supervision signal as "\textit{Before-After}" supervision. 

Nonetheless, considering that the HO's operation alters the relative position also of the \textit{wanted} elements on the belt, straightforward solutions like subtracting (or directly comparing) pairs of corresponding \textit{before} and \textit{after} images is not effective to recover the desired segmentation masks. Thus, inspired by Zerowaste \cite{bashkirova2022zerowaste}, we detailed a self-supervised learning solution in which an auxiliary classifier \(\mathcal{K}_\theta\) is trained to distinguish between \textit{before} and \textit{after} images, and Saliency Maps (SMs) of \(\mathcal{K}_\theta\) are then computed to leverage the implicit supervision provided by the HO. Specifically, the SMs for the \textit{before} class on the corresponding images highlight the most representative regions of the \textit{before} class, which essentially correspond to the unwanted objects, as these are never or rarely seen in the \textit{after} images.  Training a neural network to identify \textit{unwanted} items via this Weakly Supervised (WS) framework would enable a vision system to automatically assess the HO's performance and provide corrective feedback, effectively learning the selection task performed by the HO directly from a video stream. This novel self-supervised paradigm not only learns a specific sorting criterion but, for each new conveyor belt, it requires only collections of before/after intervention frames, with no need for manual annotations.  

To boost research in this area, our first contribution is $\text{WS}^2$ (Weakly Supervised segmentation for Waste Sorting), a comprehensive, open-access “Before-After” dataset collected in Seruso S.p.A., a real-world plastic sorting facility in Verderio (LC), Italy. Specifically, the $\text{WS}^2$ dataset comprises a sequence of more than 11 000 video frames acquired \textit{before} and \textit{after} a HO selecting semi-transparent colored PET objects from a conveyor line where only plastic items are present (Figure~\ref{fig:retain_and_remove}). This corresponds to a valuable benchmark in a waste sorting modern scenario. Unlike existing datasets (see Section~\ref{ssec:waste_images_dataset}), we specifically collected a large number of video frames to test WS segmentation methods exploiting temporal correlation in videos like \cite{marelli2025temporal}.

In our second contribution, we design a comprehensive pipeline for general SM-based WS segmentation methods to exploit the \textit{Before-After} supervision paradigm. The pipeline, detailed in Section~\ref{sec:pipeline}, encompasses SM computation, map refinement, and pseudo-mask generation to finally train a FS segmentation network without any form of human supervision. We implement our pipeline using mainstream SM methods, and we benchmark all these solution on ${\text{WS}}^2$.

Finally, we introduce a novel three-class training strategy for the auxiliary classifier $\mathcal{K}_\theta$ based on Background Removal (BR) (see Section~\ref{sec:bg_removal_prop_sol}). Since all images within the same class share a similar background, the classifier may inadvertently rely on background cues rather than object-specific features. To mitigate this bias, we explicitly model the background as a separate class alongside \textit{before} and \textit{after}, guiding the network to focus on object-level differences rather than background similarities and enhancing its ability to localize \textit{unwanted} items. Our experiments demonstrate the potential for deriving generalizable WS segmentation solutions to enable the development of deep learning-based automation for manual sorting and quality control activities.

The "Before-After" training strategy we consider was initially briefly introduced in ZeroWaste \cite{bashkirova2022zerowaste}, which focuses on high-level material segmentation (e.g., glass, cardboard, plastics) using a FS annotated dataset. ZeroWaste does not detail a pipeline to learn from the \textit{Before-After} supervision and the \textit{Before/After} data comprises a limited set of 1,400 images to separate only white paper from a mixed-material stream. This coarse formulation overlooks the finer-grained requirements of modern waste sorting scenarios, where distinctions between visually similar materials (e.g., different types of plastic) are often critical, even if humanly recognizable. Moreover, ZeroWaste does not account for background bias or temporal continuity between frames. In contrast, we show that leveraging both temporal coherence and a BR-based training strategy leads to significantly improved accuracy in WS video segmentation. 

The paper is structured as follows. Section~\ref{sec:related_work} reviews related work. Section~\ref{sec:pipeline} introduces the \textit{Before-After} training pipeline, while section~\ref{sec:bg_removal_prop_sol} focuses on the BR three-class training strategy we prepare. Section~\ref{sec:dataset} presents the key characteristics and collection methodology of our dataset. Section~\ref{sec:experiments} reports our experimental results, and Section~\ref{sec:conclusions} concludes the paper and outlines future directions.
\section{Related Work}
\label{sec:related_work}
We discuss in this section the deep learning solutions previously implemented for both human activity understanding and waste sorting management. Section~\ref{ssec:ws_human_activity} explores related work on human activity understandig in similar contexts, while Section~\ref{ssec:waste_images_dataset} illustrates existing waste datasets in the literature.

\begin{table*}[tp]
  \centering
  { \tiny
  \resizebox{\textwidth}{!}{
    \begin{tabular}{|c|c|c|c|c|c|c|c|}
      \hline
      \textbf{Dataset} & \textbf{Images} & \textbf{Task} & \textbf{Environment} & \textbf{Supervision} &\textbf{Labels} &\textbf{Labels type} & \textbf{Video} \\\hline
            TrashNet\cite{yang2016classification} (2016) & 2 400 & Classification & Laboratory & Full & 6 & Materials & No \\
            LWW \cite{sousa2019automation} (2019) & 1\,402 & Classification & In the wild & Full & 19 & Materials & No\\  
            TACO\cite{proencca2003taco} (2020) & 1\,500 & \textbf{Segmentation} & Laboratory & Full & 28 & Materials & No \\ 
            AquaTrash\cite{panwar2020aquavision} (2020) & 369 & Detection & In the wild & Full & 4 & Materials & No \\ 
            TrashCan\cite{hong2020trashcan} (2020) & 7 212 & Det./\textbf{Segm.} & In the wild & Full & 16 & Materials & \textbf{Yes} \\
            ReSortIT\cite{koskinopoulou2021robotic} (2021) & 21\,600 & \textbf{Segmentation} & Laboratory & Full & 4 & Materials & No \\ 
            Trashbox\cite{kumsetty2022trashbox} (2022) & 17\,785 & Classification & In the wild & Full & 20 & Materials & No \\  
            ZeroWaste-f \cite{bashkirova2022zerowaste} (2022) & 12\,125 & \textbf{Segmentation} & \textbf{Industrial} & Full & 4 & Materials & No \\
            ZeroWaste-w \cite{bashkirova2022zerowaste} (2022) & 2\,410 & \textbf{Segmentation} & \textbf{Industrial} & \textbf{Weak} & 2 & \textbf{Before-After} & No \\ 
            WaRP-C \cite{yudin2024hierarchical} (2024) & 10\,406 & Class./\textbf{Segm.} & \textbf{Industrial} & \textbf{Weak} & 28 & Materials & No\\  
            WS$^2$ (Ours)& \textbf{11\,060} & \textbf{Segmentation} & \textbf{Industrial} & \textbf{Weak} & 2 & \textbf{Before-After} & \textbf{Yes}\\ 
      \hline
    \end{tabular}
  }
  } 
  \caption{Summary comparison of WS² against existing waste-image datasets. For each dataset we report the number of images, the primary task (classification, detection or segmentation), the acquisition environment, the supervision regime, the number of labels and their annotation type, and whether video data are provided. Shared attributes between WS² and others are highlighted in \textbf{bold}, for a direct comparison of our contributions. WS² is the only dataset that combines \textit{Before-After} weak supervision with video sequences for segmentation in an industrial setting, and with 11 060 frames offers more than four times the images of the only other industrial  \textit{Before-After} supervised collection (ZeroWaste-w, 2 410). Moreover, the table underscores the community’s increasing emphasis on realistic, scalable benchmarks tailored to industrial environments and weakly supervised paradigms, owing to their enhanced broader applicability compared to alternative approaches.}
  \label{tab:db_table}
\end{table*}

\subsection{Human Activity Understanding}
\label{ssec:ws_human_activity}

Our work connects to the broader field of skilled human–action understanding, where scarce annotations often motivate “learning by observing” approaches. A substantial thread of research investigates procedural operations in instructional videos, aiming to recognize complex steps without dense annotation. Cross‐task weak supervision \cite{zhukov2019cross}, for instance, leverages text‐video alignments to transfer knowledge across procedures, while \cite{lin2022learning} uses narrated scripts to guide procedural activity recognition in long untrimmed clips. Egocentric video task translation \cite{xue2023egocentric} demonstrates how heterogeneous video understanding tasks can be unified by mapping outputs between multiple objectives, and \cite{jia2022egotaskqa} benchmarks comprehension of human tasks through question–answering over egocentric recordings. Although these methods excel at classifying or translating action sequences, they do not aim to train a model in reproducing or executing observed operations.

Another line of work focuses on procedural action recognition in industrial‐like settings. Several egocentric and exocentric datasets capture two‐handed assembly \cite{aganian2023attach}, anomalous manufacturing processes \cite{moriwaki2022brio}, and human–object interactions in industrial environments \cite{ragusa2021meccano}. More recently, \cite{schoonbeek2024industreal} introduces procedure step recognition with explicit modelling of execution errors in egocentric videos. While these benchmarks richly annotate object states and hand poses, they rely on full supervision and do not directly learn action dynamics purely by observing environmental changes. Finally, human-in-the-loop feedback has been applied to anomaly detection in illegal timber trade \cite{datta2023timbersleuth}, where HOs iteratively label incoming data and progressively refine an anomaly score model, in an online learning process.

In our work, we introduce a novel framework to learn not from direct observation of the human actor but from the environment they alter: by comparing the scene \textit{before} and \textit{after} each intervention, we automatically extract, segment, and understand the performed environment modification. To explore this learning framework in a real-world scenario, we collected our WS$^2$ dataset in a waste sorting facility. To our knowledge, no prior work has primarily focused on such a self-supervised learning framework.

\subsection{Waste Images Datasets}
\label{ssec:waste_images_dataset}
Waste identification datasets can be categorized based on their intended tasks: classification, object detection, and segmentation. Classification datasets \cite{yang2016classification, kumsetty2022trashbox, sousa2019automation} consist of images labeled with a single image-level category (e.g., "plastic," "glass"), while segmentation \cite{bashkirova2022zerowaste, proencca2003taco, koskinopoulou2021robotic, hong2020trashcan} and object detection \cite{fulton2020trash, panwar2020aquavision} datasets provide finer pixel-level annotations. Beyond annotation type, waste datasets differ in the environment they were collected from: in the wild (outdoor, on a floor...) \cite{kumsetty2022trashbox, sousa2019automation, proencca2003taco, hong2020trashcan, fulton2020trash, panwar2020aquavision}, or in controlled laboratory settings \cite{yang2016classification, koskinopoulou2021robotic}. Only a few datasets \cite{bashkirova2022zerowaste, yudin2024hierarchical} are collected in real waste sorting plants. Table~\ref{tab:db_table} summarizes this proprierties, comparing each dataset with ours. As can be seen from the table, there is a growing interest in both industrial and WS conditions in recent times, and the only other datasets comparable to ours are WaRP\cite{yudin2024hierarchical} and Zerowaste\cite{bashkirova2022zerowaste}.

While the WaRP \cite{yudin2024hierarchical} dataset comprises images acquired in a waste sorting plant and WS segmentation pipeline, it does not allow training models using \textit{Before-After} supervision. The weak supervision in WARP is obtained by cropping the images of the belt to single objects of different materials and training a classifier to generate SMs directly for each specific material. ZeroWaste \cite{bashkirova2022zerowaste} is the first to exploit the HO's implicit \textit{Before-After} supervision dynamic, but it contains a limited WS solutions benchmark and does not fully exploit the potential of WS methods given the limited set of 1200 images for each class. Our dataset offers significantly more images, with around 9600 images (4800 collected before and 4800 after the HO) for the training set and other 1500 images for the test set, from a modern plant addressing a more specific sorting task. Moreover, our images are collected as video frames, to promote the design of WS solutions leveraging temporal coherence like POF-CAM \cite{marelli2025temporal}, while images in \cite{bashkirova2022zerowaste} are collected only sparsely.

\begin{figure}[tp]
\centering
  \includegraphics[width=0.5\textwidth]{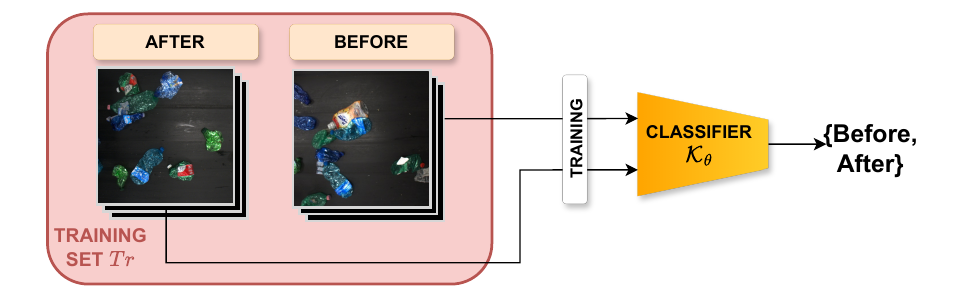}
  \caption{Overview of the first stage in our pipeline. We train the auxiliary classifier \(\mathcal{K}_\theta\) on the training set \(Tr = \{Tr^B, Tr^A\}\), to distinguish between \textit{before} (\(Tr^B\)) and \textit{after} (\(Tr^A\)) images. During the training, \(\mathcal{K}_\theta\) learns to identify discriminative features associated with \textit{unwanted} objects, since these are present only in the \textit{before} images. Once trained, \(\mathcal{K}_\theta\) is used to compute the \textit{before} saliency maps \(Sm_\textit{bef}(Tr^B)\) on the original \textit{before} training set, effectively highlighting the unwanted objects in the images.}
\label{fig:main_pipeline_training}
\end{figure}

\begin{figure*}[tp]
\centering
  \includegraphics[width=0.9\textwidth]{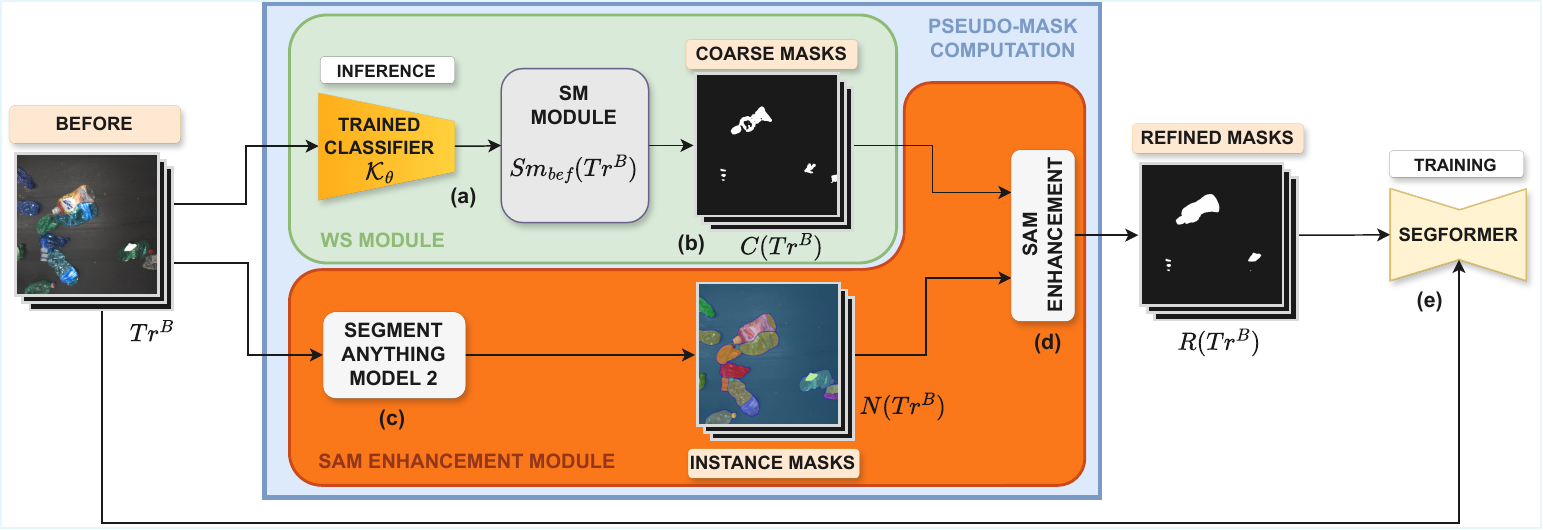}
  \caption{
    (a) In the WS module, we leverage the trained classifier \(\mathcal{K}_\theta\) to compute the saliency maps of the \textit{before} class \(Sm_\text{bef}(Tr^B)\) over the \textit{before} training set \(Tr^B\). 
    (b) We then threshold \(Sm_\text{bef}(Tr^B)\) to produce initial \textit{coarse masks} \(C(Tr^B)\).  
    (c) In parallel, in the SAM Enhancement Module, we employ SAM2 to generate \textit{instance masks} \(N(Tr^B)\). 
    (d) We then refine \(C(Tr^B)\) using the instance masks \(N(Tr^B)\), following the procedure described in~\cite{chen2023segment}, resulting in \textit{refined masks} \(R(Tr^B)\).  
    (e) Finally, we use \(R(Tr^B)\) as pseudo-mask annotations to train a FS segmentation model on $Tr^B$, SegFormer~\cite{xie2021segformer}.
  }

\label{fig:pipeline_inference}
\end{figure*}
\section{Learning with Before-After Supervision}
\label{sec:pipeline}

In this section, we detail the end-to-end pipeline we designed to learn with the \textit{Before-After} supervision framework.  Our goal is to obtain a semantic segmentation model that, given an image, either acquired \textit{before} or \textit{after} the HO, produces a binary mask accurately segmenting the \textit{unwanted} items in the image. Given the highly occluded nature of the scene, we designed our pipeline to meet two key requirements: the resulting masks must exhibit both high semantic precision, ensuring that only truly \textit{unwanted} items are identified, and fine-grained spatial accuracy, with masks boundaries closely aligning with the actual contours of the objects. The pipeline is composed of two stages, each addressing a specific task.\\\\
\textbf{Auxiliary Classifier Training:} In the first stage (Figure~\ref{fig:main_pipeline_training}), we leverage the \textit{Before-After} separation of images to train an auxiliary classifier $\mathcal{K}_\theta$ to extract the visual features related to the unwanted objects. Let \(Tr^B\) and \(Tr^A\) denote the \textit{before} and \textit{after} subsets of our training set \(Tr\), respectively. We train $\mathcal{K}_\theta$ on \(Tr\), to distinguish between images in \(Tr^B\) and \(Tr^A\). In doing so, \(\mathcal{K}_\theta\) inherently learns to identify discriminative features associated with \textit{unwanted} objects, since such objects appear only in \(Tr^B\) and represent a strong visual cue to solve the auxiliary task.\\\\
\textbf{Pseudo-Mask Computation:} In the second stage (Figure~\ref{fig:pipeline_inference}), we compute segmentation masks for the \textit{unwanted} objects by leveraging the classifier \(\mathcal{K}_\theta\) trained in the first stage. Specifically, as illustrated in the WS Module in Figure~\ref{fig:pipeline_inference}.a, we employ \(\mathcal{K}_\theta\) to generate SMs for the \textit{before} class, namely \(Sm_{\text{bef}}(Tr^B)\), on each image in the \textit{before} subset \(Tr^B\). The maps \(Sm_{\text{bef}}(Tr^B)\) highlight the most discriminative regions for the \textit{before} class, which effectively correspond to the \textit{unwanted} objects. By thresholding \(Sm_{\text{bef}}(Tr^B)\), we obtain coarse binary masks \(C(Tr^B)\) that provide initial estimates of the regions containing \textit{unwanted} items (Figure~\ref{fig:pipeline_inference}.b). To improve the quality of these masks, we introduce a two-step SAM Enhancement Module. First, in Figure~\ref{fig:pipeline_inference}.c we apply SAM2~\cite{ravi2024sam} to each image in \(Tr^B\), producing fine-grained \textit{instance masks} \(N(Tr^B)\), where each individual object is segmented separately. Then, in Figure~\ref{fig:pipeline_inference}.d, we refine the the coarse masks \(C(Tr^B)\) using the instance masks \(N(Tr^B)\) by selecting SAM segments that overlap with the saliency regions, filling holes and refining boundaries following the procedure described in~\cite{chen2023segment}. This results in high-fidelity \textit{refined masks} \(R(Tr^B)\). Finally, we use the refined masks \(R(Tr^B)\) as pseudo-mask annotations to train a FS segmentation model on the \textit{before} images \(Tr^B\) (Figure~\ref{fig:pipeline_inference}.e).

\section{Background Removal Three-Classes Training Strategy}
\label{sec:bg_removal_prop_sol}

Let ${\Lambda}$ be the set of classes  ${\Lambda}$ = \{\textit{before},\textit{ after}\} and let ${Tr}^\lambda \in \{{Tr}^B, {Tr}^A\}$ be the subset of training images labeled as $\lambda \in {\Lambda}$. 
In our framework, \textit{before} and \textit{after} frames are acquired by cameras under slightly different light conditions. As a result, images $I_i^\lambda$ of the same class set $Tr^\lambda$ share nearly identical background appearances, whereas backgrounds differ significantly between \textit{before} and \textit{after} image sets.
As shown in Figure~\ref{fig:results_with_three_classes_training_strategy}.a, this particular light condition introduces a strong bias for the auxiliary classifier $\mathcal{K}_\theta$ which, when trained directly on the full training set $Tr$ under the standard \textit{Before-After} procedure (Section~\ref{sec:pipeline}), tends to identify the background cues as distinctive features rather than the objects differences. In fact, the SM of the \textit{before} class shown in Figure~\ref{fig:results_with_three_classes_training_strategy} highlights the background reflex as relevant for the class. A classifier trained in this way becomes completely useless in any WS segmentation solution, as it fails to localize the actual \textit{unwanted} objects. To mitigate this problem, we implemented a novel three-class training strategy based on Background Removal (BR). 

\begin{figure}[tp]
\centering
      \includegraphics[width=0.5\textwidth]{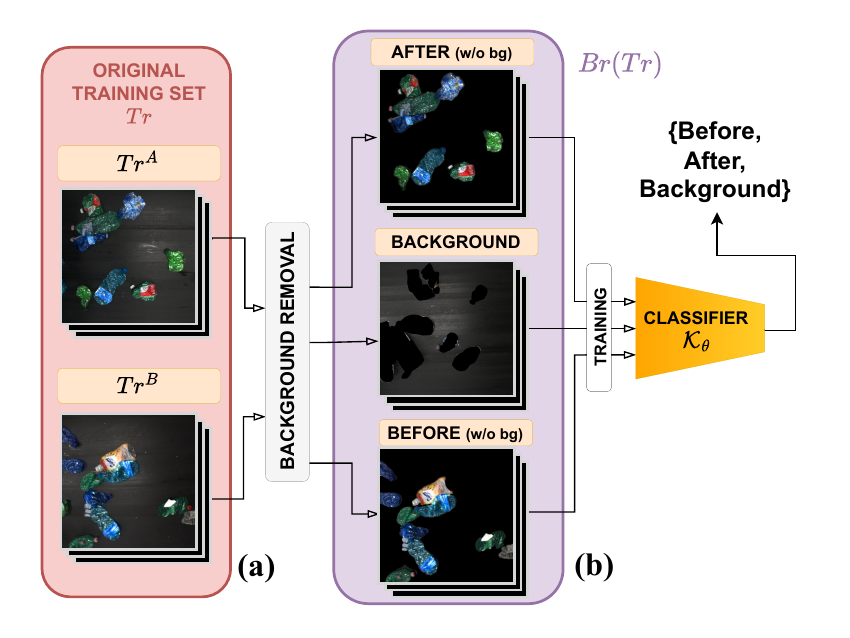}
      \caption{ The original training set $Tr$, consisting in \textit{before} ($Tr^B$) and \textit{after} ($Tr^A$) images, undergo the BR process described in Section~\ref{sec:bg_removal_prop_sol}. As output, we get a new background-removed training set $Br(Tr)$ of three classes: one containing only the extracted background images and the other two representing the \textit{before} and \textit{after} images without background. These three classes (\textit{before}, \textit{after}, and \textit{background}) are used to train the auxiliary classifier $\mathcal{K_\theta}$.}
    \label{fig:pipeline_br}
\end{figure}

\begin{figure}[tp]
\centering
    \begin{minipage}{0.35\textwidth}
        \begin{minipage}{0.45\textwidth}
            \centering
            \includegraphics[height=\textwidth]{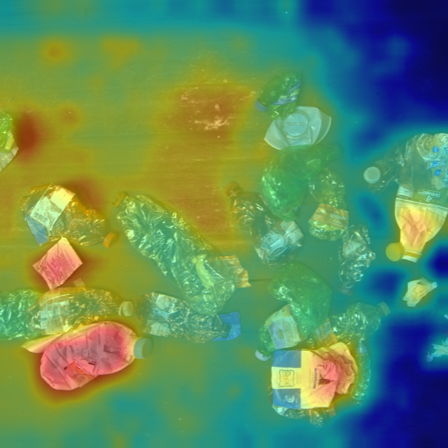}\\
            (a)
        \end{minipage}
        \hfill
        \begin{minipage}{0.45\textwidth}
            \centering
            \includegraphics[height=\textwidth]{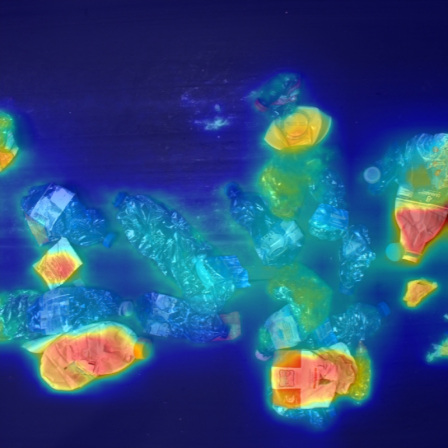}\\
             (b)
        \end{minipage}
        \vfill
        \begin{minipage}{0.45\textwidth}
            \centering
            \includegraphics[height=\textwidth]{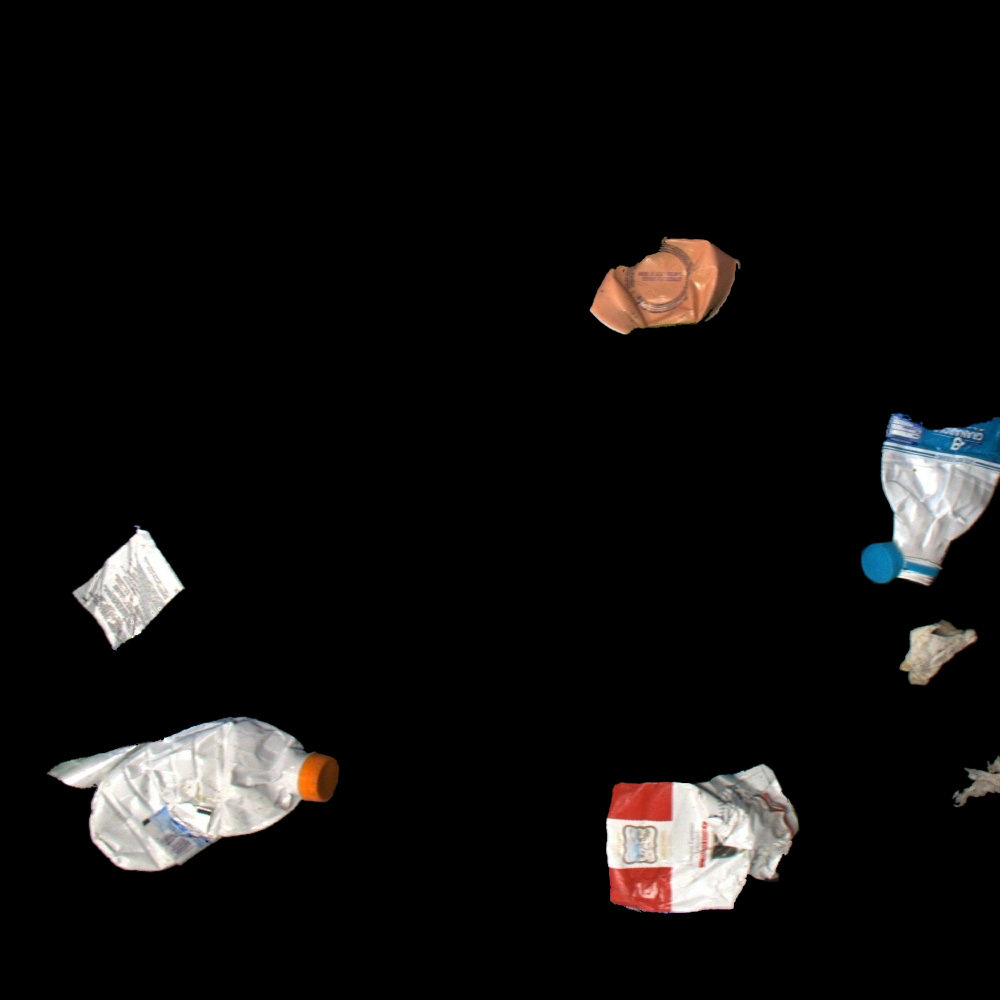}\\
            (c)
        \end{minipage}
        \hfill
        \begin{minipage}{0.45\textwidth}
            \centering
            \includegraphics[height=\textwidth]{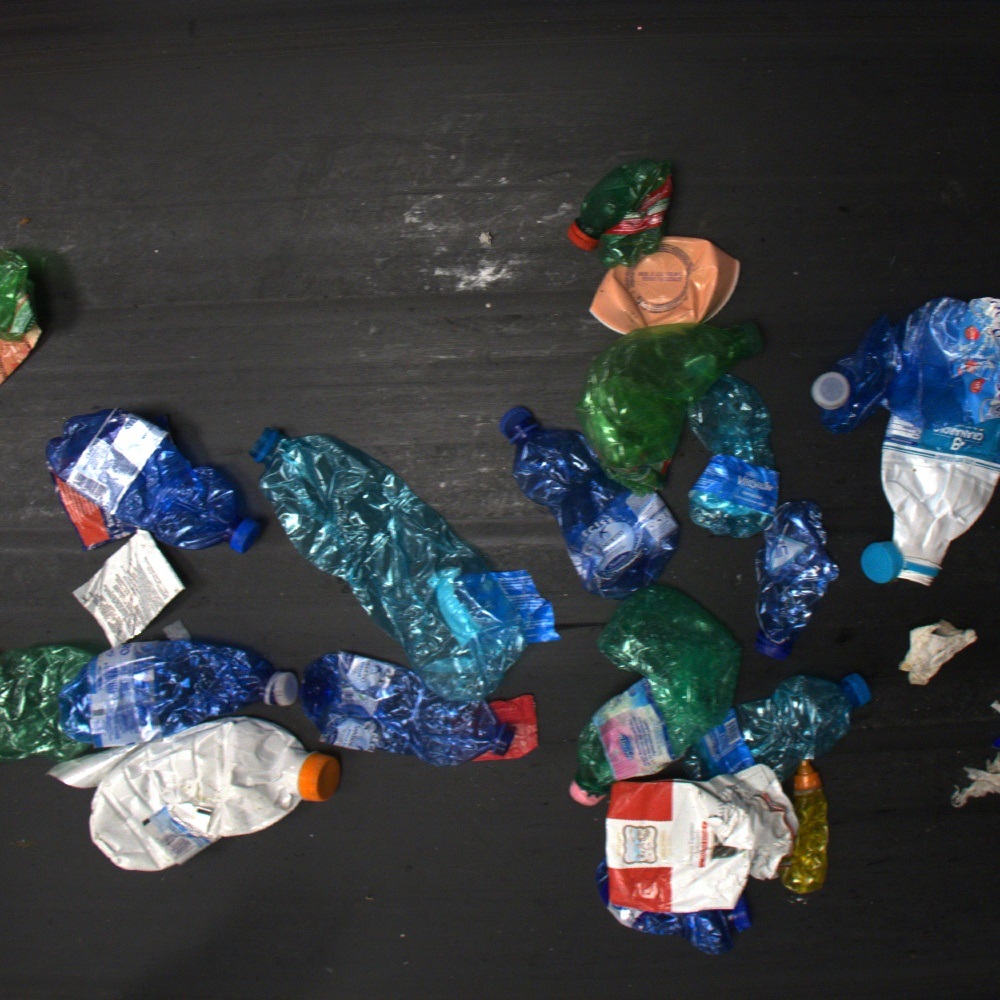}\\
            (c)
        \end{minipage}
    \end{minipage}
    
        \caption{(a) \textit{Before} image saliency map ($Sm(I^B)$) from a classifier (POF-CAM \cite{marelli2025temporal}) trained with standard \textit{before}-\textit{after} images with the background included (as in Figure~\ref{fig:main_pipeline_training}). (b) Saliency map from the same classifier trained with the BR three-classes-strategy (as in Figure~\ref{fig:pipeline_br}). (c) Ground Truth mask. (d) Original image. Training $\mathcal{K}_\theta$ on the three-class $Br(Tr)$ leads to more accurate SM to localize the \textit{unwanted} items than training it on $Tr$.}
        \label{fig:results_with_three_classes_training_strategy}
\end{figure}

To unbind the background information from class identity, we compute for each image \( I_i^\lambda \) a binary foreground/background mask \( M^{bg}(I_i^\lambda) \) using robust statistical estimation. We then use these masks to construct a new training set $Br(Tr)$ for the auxiliary classifier $\mathcal{K}_\theta$, consisting of three categories: (i) background-masked \textit{before} images, (ii) background-masked \textit{after} images, and (iii) background-only images, where foreground objects have been masked out. In this way, we expand the dataset to a third class of images consisting solely of background images, shifting from the binary set $\Lambda$ to the three-class set $\hat{\Lambda} = \{\textit{before}, \textit{ after}, \textit{ background}\}$. The resulting $Br(Tr)$, shown in Figure~\ref{fig:pipeline_br}.b, is then used to train $\mathcal{K}_\theta$ instead of the original $Tr$. As illustrated in Figure~\ref{fig:results_with_three_classes_training_strategy}.b, modeling the background as an explicit class reduces the likelihood that \( \mathcal{K}_\theta \) will rely on background features when generating saliency maps for the \textit{before} class. To maintain class balance during training, background-only images are computed from half of the \textit{before} images and half of the \textit{after} images.

To compute background binary masks $M^{bg}(I_i^\lambda)$ we rely on two assumptions:
(i) all images within the same class share an almost identical background, so foreground regions can be identified as areas that significantly deviate from the pixel-wise median image of that class;
(ii) since the background (the conveyor belt) is predominantly gray with white reflexes, pixels with high color saturation are more likely to belong to foreground objects. The detailed description of the BR procedure to compute the binary masks is detailed in Section 1 of the supplementary material. 

\section{The WS\texorpdfstring{$^2$}{2} Dataset}
\label{sec:dataset}
In this section, we introduce the WS$^2$ Dataset, specifically designed for training WS segmentation models for sorting process. The dataset captures a real-world conveyor belt where only colored semi-transparent PET objects are selected from a stream of mixed plastics, providing high-resolution video acquired both \textit{before} and \textit{after} the HO's activity. Section ~\ref{ssec: data_collection} details the acquisition campaign and setup, while Section ~\ref{ssec: WS_approach} describes the dataset's structure.

\subsection{Data Collection}
\label{ssec: data_collection}
Our dataset was acquired in the context described in Section~\ref{sec:intro}. Two cameras (Blackfly S BFS-PGE-200S6C), having resolution 5472 x 3648 pixels, were positioned to capture images \textit{before} and \textit{after} the HO (as in Figure ~\ref{fig:camera_scenario}). The cameras were fixed on ceiling-mounted supports one meter above the conveyor belt and approximately two meters apart from each other, to ensure stability and avoid vibrations of the belt or interference with the work of the HO. 

Given the high resolution and the large field of view of the cameras, it was necessary to crop images to capture only relevant sections of the conveyor belt. Images were cropped to 1000 x 1000 pixels to cover the entire belt width while excluding the HOs and their work area, which might introduce a bias when training $\mathcal{K}_\theta$. The belt moved at approximately 1 m/s, and videos were acquired at 12 fps to maintain a temporal relation between consecutive frames. Additionally, we reduced the exposure time and increased the gain to avoid motion blur and acquire sharp and bright images. Images were saved in JPEG format to optimize storage without significantly compromising their quality.

\subsection{The WS\texorpdfstring{$^2$}{2} and the Waste Sorting Task}
\label{ssec: WS_approach}
We selected a conveyor belt processing PET materials, including various types of PET: transparent, bluish, and opaque. The HOs leave only semi-transparent colored PET \textit{wanted} objects on the belt, removing all the other \textit{unwanted} objects. Consequently, the \textit{before} images capture the initial mixed material flow, while the \textit{after} images contain exclusively selected semi-transparent colored PET objects. The dataset comprises 11060 images, divided into 313 video sequences for the \textit{after} class and 284 for the \textit{before} class. Comprehensively, training and validation sets comprise 4712 \textit{after} and 4851 \textit{before} unlabeled images.

The test set comprises 1001 \textit{after} and 496 \textit{before} images, all annotated at the pixel level solely for performance assessment. Expert annotators trained in recognizing \textit{wanted} and \textit{unwanted} objects manually drew a bounding box around \textit{unwanted} objects, and let SAM (Segment Anything Model) \cite{ravi2024sam} produce segmentation maps. SAM supports diverse input prompts like points, boxes, or masks, making it adaptable to various tasks, including WS segmentation. Segmentation masks were then manually refined at a pixel level by annotators, resulting in a labeled test set for semantic segmentation. We deliberately limited manual annotations to the test set for two reasons. First, the annotation process is extremely labor-intensive and costly, making full-dataset labeling prohibitive. Second, by constraining annotations to the evaluation split, we underscore the viability of WS learning as a cost-effective alternative for training automatic recognition models.

Finally, while the \textit{after} images primarily contain semi-transparent colored PET objects, occasional anomalies may appear, as HO can sporadically overlook some items. However, although our dataset’s labels reflect human performance, such errors are random and infrequent rather than systematic. Therefore, our annotations remain sufficiently reliable as these anomalies can be treated as casual noise when training $\mathcal{K}_\theta$. Using robust deep learning models can correct these occasional mistakes and exceed human selection performance. With this dataset, we aim to foster further research into WS methods in industrial sorting.

\section{Experiments}
\label{sec:experiments}

In our experimental setup we benchmark WS$^2$ by comparing the segmentation performance of diverse SM methods based on Class Activation Maps (CAMs), focusing on the segmentation of \textit{unwanted} objects. CAMs are interpretability tools that, in the context of image classification, highlight the contribution of each pixel to a specific class \cite{zhou2016learning}. All methods are trained using the BR three-class strategy described in Section~\ref{sec:bg_removal_prop_sol}, even if testing is performed on images with the background included. This avoids the risk of discarding useful information in test images during BR. Using the same setting, all experiments with standard two-class training consistently underperformed (qualitative comparison in Figure~\ref{fig:results_with_three_classes_training_strategy}). We divded the selected tested methods in three distinct groups.  

The first group includes traditional CAM methods built upon a standard classifier, such as GradCAM,\cite{selvaraju2017grad}, GradCAM++ \cite{chattopadhay2018grad} and LayerCAM \cite{jiang2021layercam}.  These methods were selected as they do not require any special training procedure and simply operate on the auxiliary classifier $\mathcal{K}_\theta$. The main drawback is that a pixel $p$ belonging to an object of class $\lambda$, might not receive a high classification score when it does not belong to a relevant classification pattern, resulting in small segmentation masks. We trained the classifier $\mathcal{K}_\theta$ used for these methods using a categorical cross-entropy loss and an Adam optimizer at a learning rate of 5x10$^{-4}$.

The second group includes methods that incorporate additional learning constraints, such as spatial and temporal consistency, by introducing additional reconstruction losses during the classifier training to enhance the SMs \cite{jo2021puzzle, marelli2025temporal}. Specifically, PuzzleCAM \cite{jo2021puzzle} integrates a \emph{puzzle module} that divides the image into non-overlapping patches and enforces a reconstruction loss between the CAM of the full image and the combined CAMs of the patches, encouraging the global SM to retain fine-grained local details. Puzzle Optical Flow CAM (POF-CAM) \cite{marelli2025temporal} is a WS video-segmentation method that extends \cite{jo2021puzzle} by incorporating temporal information from video sequences, using optical flow between consecutive frames within the \textit{before} and \textit{after} streams to enhance the CAMs. These methods are trained according to their default setups: SGD optimizer and a multi-label soft margin classification loss. We tuned the hyperparameters for each method and, for the training losses described in \cite{jo2021puzzle, marelli2025temporal}, we set $\alpha = 2$ and $\beta = 6$. 

The third category includes the transformer-based CAM method WeakTr~\cite{zhu2023weaktr}, which first computes coarse CAMs from patch token embeddings using a convolutional layer, then refines them using self-attention maps. Both coarse and refined CAMs are jointly used for classification, enabling image-level supervision to yield detailed SMs. We replace WeakTr’s online refinement with the SAM-based refinement for a fair comparison with other methods.

We assess segmentation performances using the mean Intersection over Union (IoU), computed across various processing stages, on the annotated test set. For each method, we used a fixed 80/20 train/validation split and a maximum of 25 training epochs with early stopping to prevent overfitting. Images are resized to 512×512 and processed with a batch size of 32. We apply random color jitter augmentation to slightly modify brightness, contrast and saturation. We employed ResNet50 as backbone for all methods, except for WeakTr, which requires a transformer-based backbone, for which we used a DeIT \cite{touvron2021training}. 

\begin{figure}[tp]
\centering
\begin{minipage}{0.115\textwidth}
    \centering
    \includegraphics[height=2cm]{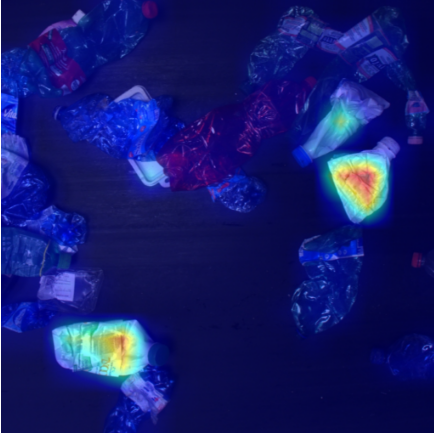}\\
    GradCAM
\end{minipage}
\hfill
\begin{minipage}{0.115\textwidth}
    \centering
    \includegraphics[height=2cm]{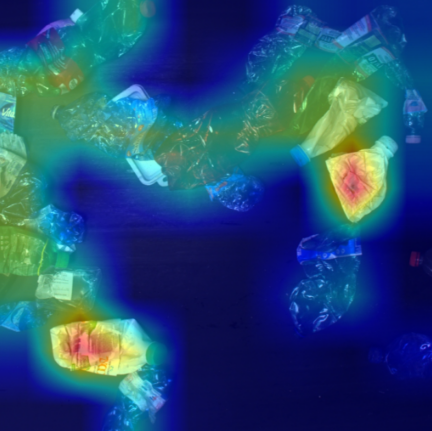}\\
    GradCAM++
\end{minipage}
\hfill
\begin{minipage}{0.115\textwidth}
    \centering
    \includegraphics[height=2cm]{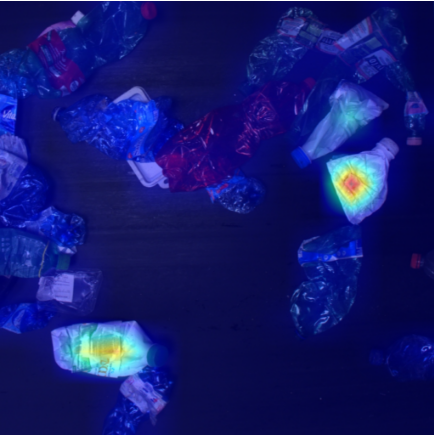}\\
    LayerCAM
\end{minipage}
\hfill
\begin{minipage}{0.115\textwidth}
    \centering
    \includegraphics[height=2cm]{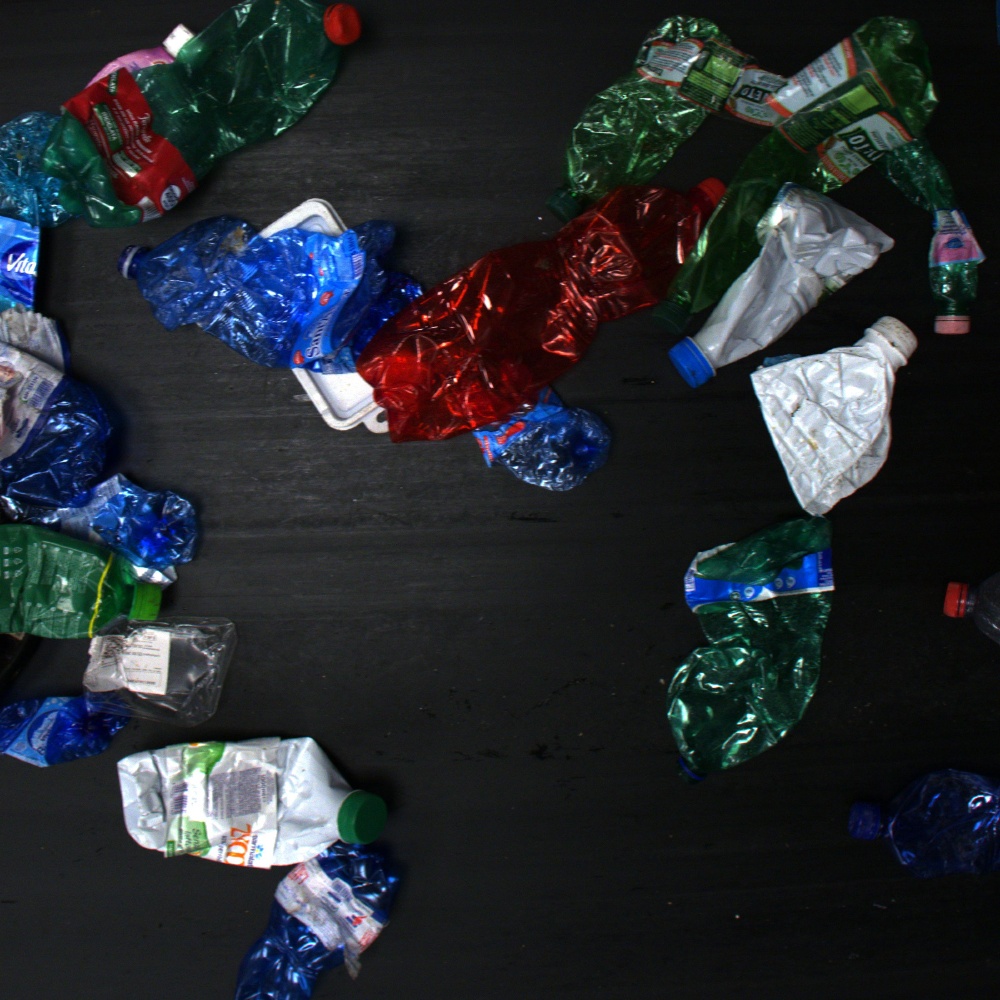}\\
    Image
\end{minipage}
\vspace{0.2cm}
\vfill
\begin{minipage}{0.115\textwidth}
    \centering
    \includegraphics[height=2cm]{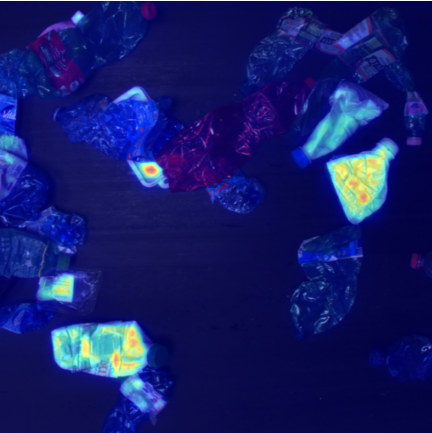}\\
    PuzzleCAM
\end{minipage}
\hfill
\begin{minipage}{0.115\textwidth}
    \centering
    \includegraphics[height=2cm]{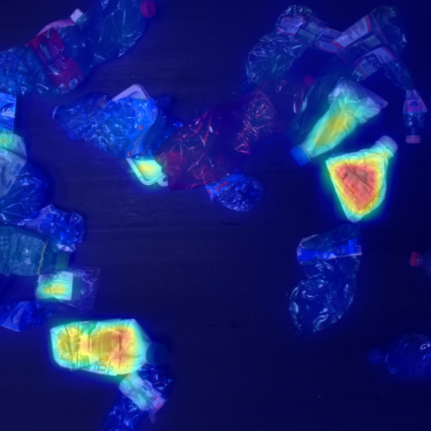}\\
    POF-CAM
\end{minipage}
\hfill
\begin{minipage}{0.115\textwidth}
    \centering
    \includegraphics[height=2cm]{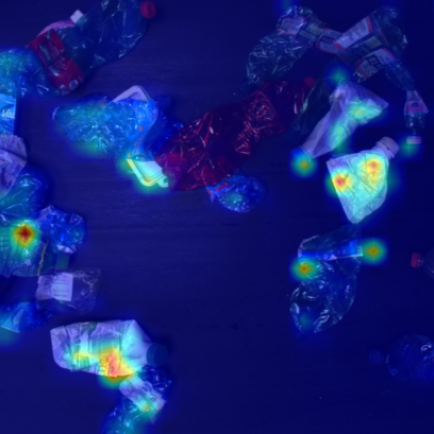}\\
    WeakTr
\end{minipage}
\hfill
\begin{minipage}{0.115\textwidth}
    \centering
    \includegraphics[height=2cm]{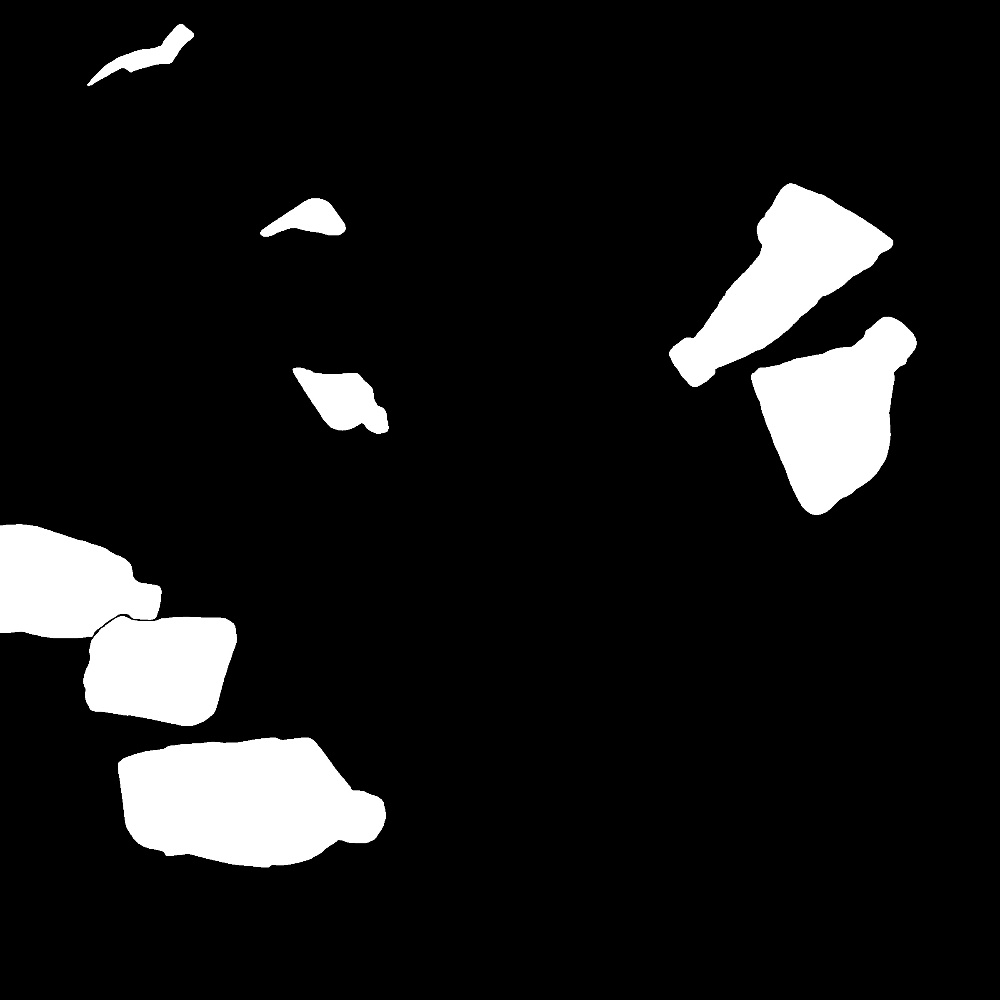}\\
    Ground Truth
\end{minipage}
\caption{Qualitative examples of SMs obtained using different WS methods on an image taken from the test set compared with the original RGB image and the ground truth mask.}
\label{fig:cams_of_different_types}
\end{figure}

To ensure consistency and reliability, all experiments focus on segmenting \textit{unwanted} objects in the same test set $Ts$, which is never used for training or validation at any stage. We assess the accuracy of the segmentation masks returned at different levels of our pipeline for each method on both \textit{before} and \textit{after} images of $Ts$, namely $Ts^B$ and $Ts^A$, to highlight the performance difference for each class. The three distinct levels of our benchmark of the state-of-the-art are defined as follows:

\begin{enumerate}[label=\textbf{(\roman*)}]
  \item \textbf{Coarse Masks \(\mathbf{C(Ts)}\) mIoU}: We measure the mIoU directly on the coarse masks \(C(Ts)\) derived by the \textit{before} CAM \(M_{bef}(Tst)\) generated by the classifier \(\mathcal{K_\theta}\) on both \(Tst^A,\,Tst^B \subset Tst\).
  
  \item \textbf{Refined Masks \(\mathbf{R(Ts)}\) mIoU}: We refine the coarse masks \(C(Tst)\) using the Sam Enhancement Module illustrated in Figure~\ref{fig:pipeline_inference}.b. Instance masks \(N(Tst)\) are computed by SAM2 \cite{ravi2024sam} and used by the algorithm described in \cite{chen2023segment} to obtain the refined masks \(R(Tst)\). The mIoU for this stage is measured on \(R(Tst)\).
  
  \item \textbf{SegFormer Masks \(\mathbf{S(Ts)}\) mIoU}: We measure the mIoU of the segmentation masks \(S(Tst)\) returned from SegFormer \cite{xie2021segformer} trained on \(Tr^B\) and \(R(Tr^B)\).
\end{enumerate}
Table~\ref{tab:miou_table} reports the mIoU scores at each evaluation stage, while Figure~\ref{fig:cams_of_different_types} shows qualitative examples of saliency maps produced by different WS methods before thresholding into \( C(Ts) \), alongside the corresponding original images and ground-truth masks.

\begin{table}[ht!]
\centering
\resizebox{\columnwidth}{!}{
\begin{tabular}{|c|cc|cc|cc|}
\hline
\textbf{Models} & \multicolumn{2}{c|}{\textbf{$C(Ts)$}} & \multicolumn{2}{c|}{\textbf{$R(Ts)$}} & \multicolumn{2}{c|}{\textbf{$S(Ts)$}} \\
\cline{2-7}
                & \textbf{$Ts^B$} & \textbf{$Ts^A$} & \textbf{$Ts^B$} & \textbf{$Ts^A$} & \textbf{$Ts^B$} & \textbf{$Ts^A$} \\
\hline
GradCAM \cite{selvaraju2017grad}     & 19.86 & 8.95  & 25.03 & 12.67 & 27.64 & 14.10 \\ 
GradCAM++ \cite{chattopadhay2018grad}   & 16.40 & 6.54  & 20.08 & 9.94  & 25.74 & 7.72  \\ 
LayerCAM \cite{jiang2021layercam}    & 17.69 & 9.3   & 21.31 & 12.70 & 28.10 & 14.10 \\ 
PuzzleCAM \cite{jo2021puzzle}   & 33.52 & 15.29 & 35.35 & 16.99 & 37.15 & 13.74 \\ 
POF-CAM \cite{marelli2025temporal} & \textbf{38.70} & \textbf{20.93} & \textbf{41.40} & \textbf{23.08} & \textbf{42.58} & \textbf{19.78} \\ 
WeakTr \cite{zhu2023weaktr}& 21.44 & 8.20  & 23.73 & 9.62  & 24.14 & 4.42  \\ 
\hline
\end{tabular}
}
\caption{Mean Intersection Over Union (mIoU) percentage (\%) of SOTA on \textit{before} and \textit{after} Images and in three stages: raw CAM, after SAM-Refinement and after SegFormer. POF-CAM\cite{marelli2025temporal}, using temporal consistency, significantly outperforms other methods.}
\label{tab:miou_table}
\end{table}

Since WS methods identify the regions of an image \(I\) that are most relevant for $\mathcal{K}_\theta$ to assign \(I\) to a specific class, and since the \textit{unwanted} object represents the key region for the \textit{before} class, segmentation masks in \textit{after} images leads to poor performance. This happens because SMs attempt to identify \textit{before} features in images that belong to a different class. This performance gap between the two classes persists even in the masks estimated in the next refinement steps. Furthermore, the resulting class imbalance and the limited accuracy of pseudo-masks can induce negative learning in the SegFormer network, which can't effectively disentangle the concept of \textit{unwanted} objects from the \textit{before} class, leading to inconsistent generalization. As a result, while segmentation performance steadily improves on \textit{before} images, it may degrade on \textit{after} ones.

Another remarkable piece of evidence is the impact of the SAM refinement, improving the performance of each method by several mIoU points. This shows that the level of detail brought by vision foundation models remains difficult to attain by WS segmentation alone methods alone. Nonetheless, the choice of the WS model remains crucial. Table~\ref{tab:miou_table} shows that the ranking of the best-performing models on our task remains very similar after refinement, consolidating the fact that, even with refinement, highly detailed SMs are needed to obtain good masks using SAM, as some methods perform better than others at localizing semantically relevant regions. This becomes especially true in use cases like waste sorting, where we can appreciate a considerable distribution shift with respect to the datasets used to train foundation models used for refinement.

Remarkably, we can observe how the methods employing additional learning constraints \cite{jo2021puzzle, marelli2025temporal} outperform the others. In particular, the spatial consistency provided by the puzzle module helps to detect multiple objects of the same class scattered across the image, which is the standard setting in our dataset. POF-CAM  \cite{marelli2025temporal}, the only method that leverages temporal consistency, significantly outperforms all other solutions, highlighting the critical role of temporal information in a dataset of this nature, a factor absent in previous works. Notably, WeakTr, despite being state-of-the-art in CAM generation, performs poorly on our dataset compared to methods employing puzzle modules. This highlights its limited generalization capability for specific tasks relative to convolutional-based approaches.
\section{Conclusions}
\label{sec:conclusions}

We introduced WS$^2$, the first large‐scale, multiview video dataset for weakly supervised segmentation in industrial waste sorting. Our dataset captures over 11 000 “before” and “after” video frames around a manual removal operation on a plastic processing conveyor belt. To take advantage of the \textit{Before-After} supervision implicit in the human removal operation, we developed a three-stage pipeline in which we (i) train an auxiliary classifier to localize \textit{unwanted} items via saliency maps, (ii) refine these maps using SAM, and (iii) use the resulting refined masks to train a fully supervised segmentation network, whithout the need of any manual annotation. To mitigate the background bias intrinsic in this problem we also implemented an innovative background removal-based three-class training strategy. Our extensive benchmarks, spanning saliency maps methods of different categories, demonstrate the critical roles of temporal continuity and background bias mitigation in improving weakly supervised segmentation accuracy for this task. Our work not only eases the development of better saliency map solutions but also opens avenues for integrating temporal coherence in video-based weak supervision and skilled activity understanding across diverse domains.
{
    \small
    \bibliographystyle{ieeenat_fullname}
    \bibliography{main}
}

\end{document}